\documentclass[fleqn,10pt]{wlscirep}
\usepackage[utf8]{inputenc}
\usepackage[T1]{fontenc}

\usepackage{graphicx}%
\usepackage{multirow}%
\usepackage{amsmath,amssymb,amsfonts}%
\usepackage{amsthm}%
\usepackage{mathrsfs}%
\usepackage[title]{appendix}%
\usepackage{xcolor}%
\usepackage{textcomp}%
\usepackage{manyfoot}%
\usepackage{booktabs}%
\usepackage{algorithm}%
\usepackage{algorithmicx}%
\usepackage{algpseudocode}%
\usepackage{listings}%
\usepackage[sort]{natbib}
\usepackage{hyperref}
\usepackage{tikz,xcolor}

\setcitestyle{square}
\setcitestyle{numbers}

\title{Symbolically Regressing Fish Biomass Spectral Data: A Linear Genetic Programming Method with Tunable Primitives}

\author[1,*]{Zhixing Huang}
\author[1]{Bing Xue}
\author[1]{Mengjie Zhang}
\author[2]{Jeremy S. Ronney}
\author[3]{Keith C. Gordon}
\author[4]{Daniel P. Killeen}
\affil[1]{the Centre for Data Science and Artificial Intelligence \& School of Engineering and Computer Science, Victoria University of Wellington, Wellington, 6140, New Zealand}
\affil[2]{Department of Chemistry, University of Otago, Dunedin, New Zealand}

\affil[3]{MacDiarmid Institute for Advanced Materials and Nanotechnology, Chemistry Department, University of Otago, Dunedin, New Zealand}

\affil[4]{The New Zealand Institute for Plant and Food Research Limited, Nelson, New Zealand}

\affil[*]{zhixing.huang@ecs.vuw.ac.nz}

\begin{abstract}
Machine learning techniques play an important role in analyzing spectral data. The spectral data of fish biomass is useful in fish production, as it carries many important chemistry properties of fish meat. However, it is challenging for existing machine learning techniques to comprehensively discover hidden patterns from fish biomass spectral data since the spectral data often have a lot of noises while the training data are quite limited. To better analyze fish biomass spectral data, this paper models it as a symbolic regression problem and solves it by a linear genetic programming method with newly proposed tunable primitives. In the symbolic regression problem, linear genetic programming automatically synthesizes regression models based on the given primitives and training data. The tunable primitives further improve the approximation ability of the regression models by tuning their inherent coefficients. Our empirical results over ten fish biomass targets show that the proposed method improves the overall performance of fish biomass composition prediction. The synthesized regression models are compact and have good interpretability, which allow us to highlight useful features over the spectrum. Our further investigation also verifies the good generality of the proposed method across various spectral data treatments and other symbolic regression problems.
\end{abstract}
\begin{document}

\flushbottom
\maketitle
\thispagestyle{empty}

\section{Introduction}\label{sec1}

% in aquatic food (i.e., biomass composition) is an essential step in fish production lines.
% This paper takes fish biomass prediction based on Raman spectral data as a case study. The fish biomass prediction has a high commercial and academic potential.
% From a commercial view, aquatic food plays an important role in global food security and livelihoods. According to the FAO's report in 2024 \cite{fao_state_2024}, the global production of aquatic animals reached about 185 million tonnes in 2022, with an annual increase rate of about 3.2\% and an approximate total first sale value of USD 452 billion.
% Aquatic food is an essential source of a wide range of nutrients as well, particularly mineral and fatty acids.
% From an academic view, measuring the biomass of nutrients in fish based on spectral data is still a challenging task  \cite{zhou_machine_2024, aziz_automated_2022}. In aquatic food production, we can only scan the fish meat by laser for a short time (e.g., less than 4 minutes) to avoid heating fish meat \cite{herrero_raman_2008, qu_raman_2022, ahmmed_detection_2023}. This inevitably brings noises into fish spectral data. 

Aquatic food plays an important role in global food security and livelihoods. According to the FAO's report in 2024 \cite{fao_state_2024}, the global production of aquatic animals reached about 185 million tonnes in 2022, with an annual increase rate of about 3.2\% and an approximate total first sale value of USD 452 billion.
Aquatic food is an essential source of a wide range of nutrients as well, particularly mineral and fatty acids.

Predicting biomass compositions by spectroscopic techniques (such as infrared and Raman spectroscopy) is an important step in aquatic food production lines since it estimates nutrition compositions efficiently without destructing aquatic food \cite{xue_identification_2022, song_multimodal_2024, gade_-beam_2024}. 
% The spectroscopic technique (such as infrared and Raman spectroscopy) is an effective, fast, and non-destructive method for analyzing material composition and has been applied to a wide range of domains \cite{xue_identification_2022, song_multimodal_2024, gade_-beam_2024}.
The spectroscopic techniques usually work with machine learning methods, such as partial least square regression (PLSR) \cite{usman_soc_2023} and convolution neural networks \cite{huang_rapid_2023}, to predict biomass compositions. 
However, applying existing machine learning methods for analyzing fish biomass composition in real-world production lines is still a challenging task \cite{zhou_machine_2024, aziz_automated_2022}.
There are two main challenges. First, the data quality and quantity are very limited. The fish spectral data often have a lot of noises since we can only scan the fish meat by laser for a short time (e.g., less than 4 minutes) to avoid heating fish meat \cite{herrero_raman_2008, qu_raman_2022, ahmmed_detection_2023}.
The amount of data is insufficient because measuring the accurate modules of materials is costly and cumbersome. These likely result in overfitting in existing machine learning methods. 
% In addition, the spectroscopic data is high dimensional and usually contains noises, 
% since we only scan fish meat for a short time. If we scan fish meat with laser lights for a long time (longer than 30 minutes), the laser might heat fish meat and cook it.
Second, some existing machine learning models lack interpretability to assist human scientists in understanding spectroscopic data. Because many machine learning models are black-box models and might include millions of parameters, it is non-trivial to understand and reason the predictions from these machine learning models.

% Spectra data is essentially a certain wavelength range of reflecting lights when we hit fish meat by laser.
% Different biomass compositions likely have different spectra curves. 

% to obtain stable spectra, we have to scan fish meat with laser lights for a long time (longer than 30 minutes). However, the laser might heat fish meat and cook it. To keep fish meat fresh, we only scan fish meat for a short time, leading to unstable spectra. 
% In addition, we usually analyze the spectra data dependent on Fourier transform systems. However, the Fourier transformation has a high computation complexity, preventing real-time fish biomass analyses. 

In response to the outlined challenges in the prediction of biomass compositions, this paper proposes a linear genetic programming method with tunable primitives (denoted as LGP-TP) to automatically synthesize regression models for analyzing spectral data, i.e., regarding the spectral data analysis as a symbolic regression task. 
% In particular, we take fish biomass composition prediction based on Raman spectroscopy as a case study.
Linear genetic programming (LGP) has shown a promising learning ability and has been successfully applied to symbolic regression problems \cite{Brameier2007,Huang2022SLGP}, classification \cite{Fogelberg2005, Sotto2022Ana}, and combinatorial optimization \cite{huang_toward_2024}. 
Compared with other machine learning methods, LGP can automatically find concise models for describing knowledge in the given data \cite{winkler_how_2024}. This relieves the impact caused by the limited training data and the noises by avoiding over complicated models. LGP outputs regression models with good interpretability \cite{mei_explainable_2022}, highlighting interesting regions of spectroscopic data. It helps human scientists understand which features (or peaks in the spectrums) are useful for prediction.
In addition, LGP programs naturally reuse building blocks (i.e., sub-programs) to design compact solutions and have a similar form to human written programs (instruction-by-instruction representation). This allows LGP to develop sophisticated programs, such as harnessing control flow operators (e.g., IF and FOR-loop) \cite{Brameier2007, huang_toward_2024} and having multiple outputs \cite{Brameier2007,Huang2023MLSI}.
% In this paper, we apply LGP to synthesize regression models from scratch for fish spectroscopic data, 
% The synthesized regression model predicts biomass composition for unseen spectroscopic data.    

However, existing LGP methods usually synthesize regression models based on predefined primitives (e.g., $+$, $\times$, and $x_1$). The fixed primitives might limit the approximation ability of LGP (e.g., representing $\sin(3x+\pi)$ by given $\sin(\cdot)$ only is uneasy). To enhance LGP performance in spectral data analysis, we propose three kinds of tunable primitives that perform residual error approximation in LGP programs. The tunable primitives can represent a wide range of function mappings by tuning their parameters.
% The main idea for improving LGP performance in biomass prediction is to minimize the predicting error of LGP programs by tuning parameterized primitives.
Specifically, the three kinds of tunable primitives are 1) tunable terminals that accept input features and approximate the target outputs; 2) tunable functions that accept intermediate results of LGP programs and approximate the target outputs; 3) a multivariate linear regression function that approximates the final values of multiple registers to the target outputs.

\begin{figure*}
    \centering
    \includegraphics[scale=0.65, viewport=15 15 550 180, clip=true]{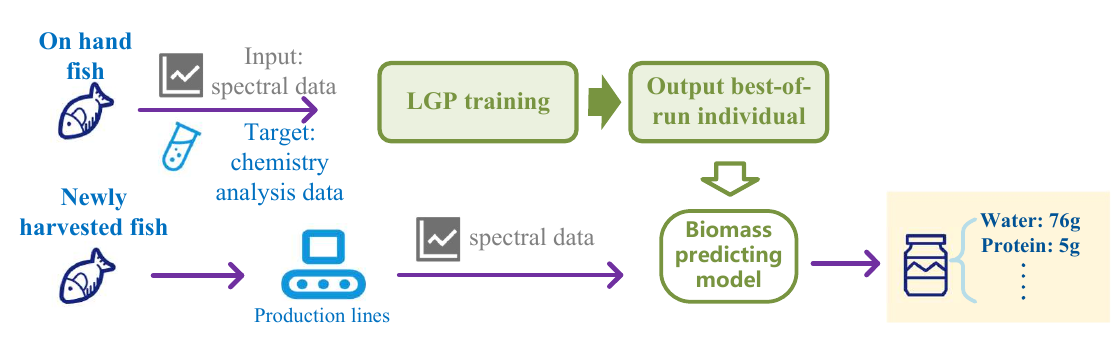}
    \caption{Schematic diagram of applying LGP-TP for fish biomass prediction. Particularly, LGP-TP synthesizes regression models based on the spectral data (inputs) and the chemical ground truth of on-hand fish samples (target outputs). The synthesized regression model from LGP-TP predicts the biomass of unseen fish samples in real-world production.}
    \label{fig:schematic}
\end{figure*}

Fig. \ref{fig:schematic} is the schematic diagram of applying LGP-TP to analyze fish biomass spectral data. 
First, we apply spectroscopic techniques and chemical analysis techniques to obtain the input and the ground truth target data of fish samples, respectively. Then, we train LGP-TP to synthesize regression models to predict the target data based on the inputs. 
After training, LGP-TP outputs the best regression models. The regression model predicts the biomass compositions of newly harvested fish samples on the production lines based on their spectral data. By this means, we can have a fast and cheap estimation to biomass composition and get rid of the expensive chemical analysis.

% $$\omega = \arg \min_{\omega}\sum^m_i(\tau(x_i, \omega) - y_i)^2$$
% $$\omega = \arg \min_{\omega}\sum^m_i(\psi(R_{j,i}, \omega) - y_i)^2$$
% $$\omega = \arg \min_{\omega}\sum^m_i(\omega \cdot [1, R_i]^T - y_i)^2$$

% The main idea for improving LGP performance in biomass prediction is to reduce the semantic differences between the input and output semantics of (sub) LGP programs by tuning parameterized components. We propose three tunable components for LGP programs. 
% First, we design a set of tunable terminals that accept input features and approximate the target outputs. 
% % for LGP and tune the parameters of the trainable primitives during evolution. 
% Second, we design a set of tunable functions that accept intermediate results of LGP programs and approximate the target outputs.
% Third, we append a multi-variable linear regression model (i.e., a special tunable function) after every LGP program. The multi-variable linear regression model approximates the multiple outputs from an LGP program to the target outputs.

The results show that LGP-TP achieves the best overall predicting performance over ten fish biomass targets, compared with six popular regression methods. The six compared methods include one common regression method for spectral data, PLSR, and one stat-of-the-art method tailored toward fish biomass prediction.
The statistical analysis confirms that LGP-TP has a significantly better prediction performance over nearly half of the targets and has a better overall performance than the other methods.
In addition, the synthesized regression models by LGP-TP have good interpretability. The synthesized models highlight the linear correlation between inputs and targets and specific peaks over the spectral data.
Finally, our further investigation of other data treatments of fish spectral data and of a popular symbolic regression benchmark, SRBench, shows its potential generality and high training efficiency.

% \subsection{Linear Genetic Programming}

% This report applies linear genetic programming (LGP) to perform symbolic regression over the spectroscopy data.
% LGP is an evolutionary computation method that searches symbolic solutions \cite{brameier_linear_2007}. LGP has many advantages over other GP methods. For example, LGP programs naturally reuse building blocks (i.e., sub-programs) to design compact solutions and have multiple outputs. In addition, LGP easily harnesses control flow operators (e.g., IF and FOR-loop) because of its instruction-by-instruction representation.

\section{Results}\label{sec2}

\subsection{Fish Spectroscopic Dataset}\label{sec2:dataset}
To verify the performance of the proposed method, this paper applies LGP-TP to predict 10 kinds of biomass based on the spectroscopic data from 39 unique fish samples of two deep-water fish species in New Zealand, Hoki and Mackerel. To reduce measuring bias, each fish sample is measured three times, resulting in a total of 117 spectroscopic instances.
The biomass targets include water, ash, nitrogen, protein, fat, lipids yield, wax/sterol esters, triacylglycerols, free fatty acids, and sterols. 
We apply InGaAs Raman at 1064nm to obtain the spectroscopic data and truncate the wavenumber range of 580.109 to 202.533 (cm$^{-1}$) to reduce the impact by glass vials of fish meat. As a result, the obtained spectroscopic data has a wavenumber range of 1891.58-580.109 (cm$^{-1}$), with 427 input features. We correct the obtained spectroscopic data by standard normal variate transformation (SNV). 

Considering the limited spectroscopic instances in the dataset, we apply six-fold cross-validation to evaluate machine learning models. Specifically, we split the 117 instances into six folds, each fold with approximately 20 instances (6 or 7 fish samples). In one six-fold cross-validation, a machine learning model alternatively takes one fold of split instances as unseen test data and trains on the other five folds, resulting in six different training (and test) performances. The mean of the six different training (and test) performances is regarded as the performance of one six-fold cross-validation. Each machine learning model runs the six-fold cross-validation for 10 times with unique random seeds.
To enhance the performance of machine learning models, we augment the training data of machine learning models 50 times (i.e., a total of approximately 5000 training instances) by three strategies: spectral data augmentation \cite{bjerrum_data_2017}, mix-up (i.e., linear interpolation) \cite{zhang_mixup_2018}, and Gaussian noise. These three strategies produce 50\%, 25\%, and 25\% of new instances, respectively.

\subsection{Comparison Design}
We compare LGP-TP with six popular machine learning methods for predicting fish biomass, including partial least squares regression (PLSR) \cite{wold_pls-regression_2001}, K-nearest neighbor (KNN), multi-layer perception (MLP), gradient boosting regression (XGB), random forest regression (RF), and FishCNN \cite{zhou_machine_2024}. Particularly, PLSR is one of the commonly used machine learning methods for analyzing spectroscopic data, and FishCNN is one of the state-of-the-art methods tailored toward fish biomass prediction based on Raman spectral data. All seven methods run 10 times six-fold cross-validation for each biomass target, respectively.

% The parameters of all the compared methods are summarized in the supplement. 
\begin{table}[t!]
\centering
\caption{Key parameters of the compared methods}
\label{tb:parameters}
\scalebox{0.85}{
\begin{tabular}{c|p{90mm}}
\hline
methods      & \multicolumn{1}{c}{key parameters}                                                 \\ \hline
PLSR         & n\_components=10                                                                   \\ \hline
KNN          & K=3 \\\hline
MLP          & max\_iter=1000, solver=`adam',activation=`Relu', alpha=0.0001                      \\ \hline
XGB          & max\_depth=5, n\_estimators=100                                                    \\ \hline
RF & max\_depth=5, n\_estimators=100                                                    \\ \hline
% EvoForest    & max\_height=5,   select='AutomaticLexicase', boost\_size=100, n\_gen=50,n\_pop=200 \\ \hline
FishCNN      & max epochs=1500, optimizer=`AdamW', 2$\times$ Convolutional layers (16 filters, 64 filter size, 1 stride), fully connected layers=[128,16,3] \\ \hline
LGP-TP          & popsize=250, n\_gen=100, max\_programsize=50, n\_registers=30                      \\\hline
\end{tabular}}
\end{table}

The parameters of all the compared methods are summarized in table \ref{tb:parameters}. 
The parameters of PLSR and FishCNN follow their recommended settings in \cite{zhou_machine_2024}, and the parameters of the other methods are the default settings.
Specifically, PLSR has 10 components, KNN has $k=3$, MLP, whose activation units are ``Relu'', learns with a solver of ``adam'' and with a learning rate of $1e^{-4}$ for 1000 iterations, and XGB and RF have up to 100 decision trees (``n\_estimators''), each with a maximum depth of 5.
FishCNN is trained by weighted adam (``AdamW'') for up to 1500 epochs. FishCNN has two convolutional layers, each with the same architecture, and three fully connected layers.
The parameters of LGP-TP mostly follow the existing studies \cite{huang_toward_2024}.
LGP-TP evolves a population of 250 individuals for 100 generations. Each LGP-TP individual has up to 50 instructions with 30 available registers.

\subsection{Prediction Performance}\label{sec2:testperformance}
\begin{figure}
    \centering
    \includegraphics[scale=0.72, viewport=20 50 560 800, clip=true]{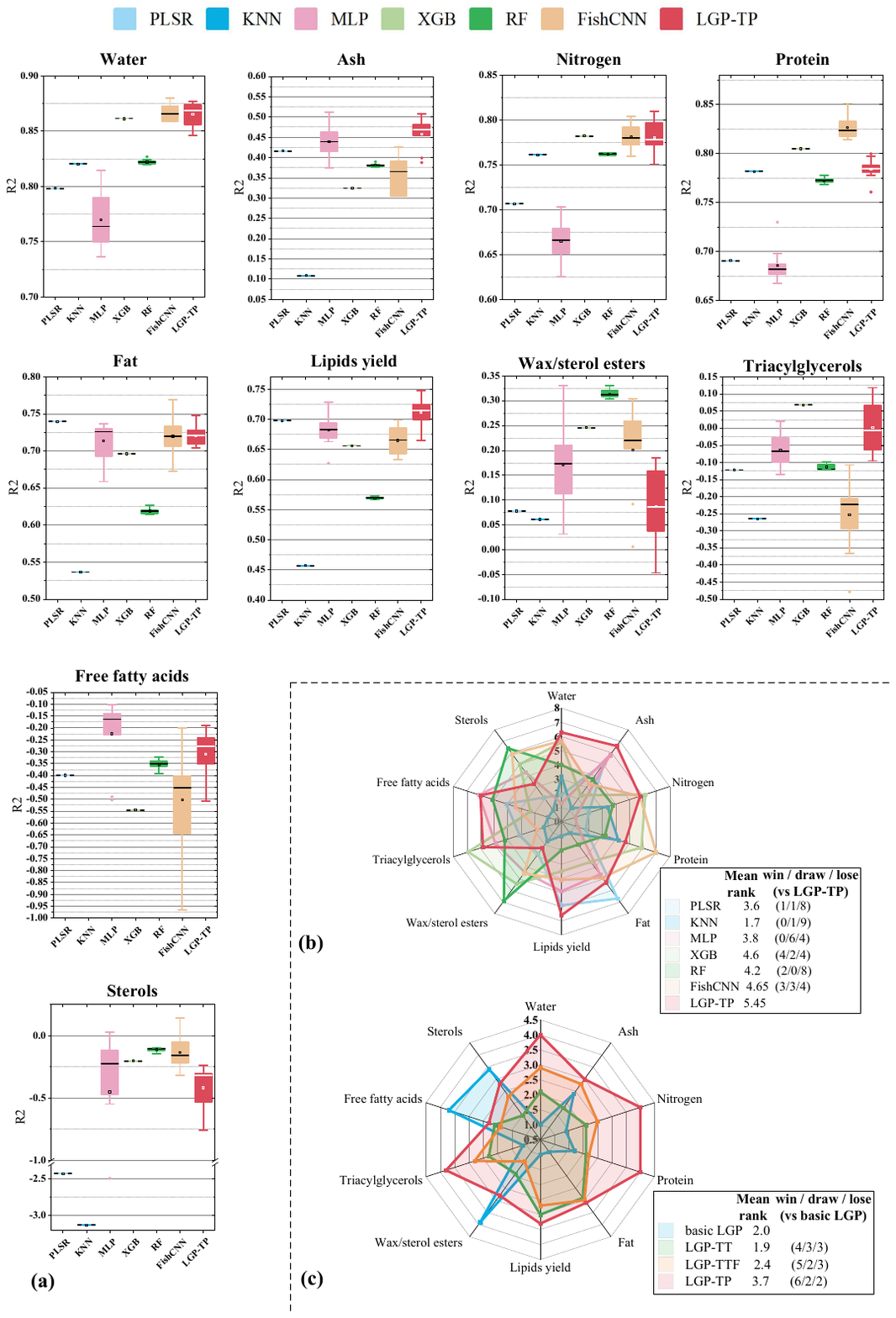}
    \caption{Test performance of the compared methods. \textbf{(a)} Test R$^2$ of the compared methods in the 10 six-fold cross-validations, each box plot for a biomass target. \textbf{(b)} The radar chart shows the mean ranks of the compared methods over different biomass targets. The legend of the chart shows the overall mean ranks by Friedmant's test and the statistical results by the Wilcoxon rank-sum test ($\alpha=0.05$). ``win'' indicates the number of targets where a compared method is significantly better than the proposed LGP-TP. ``draw'' indicates the number of targets without significantly different test R$^2$. ``lose'' indicates the number of targets where a compared method is significantly worse than LGP-TP. \textbf{(c)} The ablation study of LGP-TP by adding tunable primitives one-by-one in LGP evolution. The legend follows the same design as sub-figure b.}
    \label{fig:testR2}
\end{figure}

The main goal of this paper is to automatically construct effective regression models by LGP-TP for predicting fish biomass for newly harvested fish.
We evaluate the prediction performance based on the R$^2$ values over test instances. Fig. \ref{fig:testR2}a shows the distributions of test R$^2$ over the 10 six-fold cross-validations of the compared methods. We can see that the proposed LGP-TP has the highest test R$^2$ distributions in the biomass targets of water, ash, and lipids yield, and has a very competitive performance with other compared methods in nitrogen, fat, and triacylglycerol. To statistically analyze the performance of LGP-TP, we apply Friedman's test with a significance level of 0.05 and with the Bonferroni correction and the Wilcoxon rank-sum test with a significance level of 0.05 to the test R$^2$ over the 10 six-fold cross-validation. The Friedman test has a p-value of 0.0061, indicating a significant difference among the compared methods.
The radar chart in Fig. \ref{fig:testR2}b shows the mean ranks of the compared methods by Friedman's test over the ten biomass targets. LGP-TP (the red polygon) covers the largest area on the radar chart. Specifically, LGP-TP has the best mean ranks in the biomass targets such as water, lipids yield, and free fatty acids. The legend in Fig. \ref{fig:testR2}b summarizes the average of the mean ranks and the results of the Wilcoxon rank-sum test. We can see that the other six compared methods perform significantly worse than LGP-TP on at least four biomass targets. 
Specifically, the second-best and third-best compared methods (i.e., FishCNN and XGB) are significantly worse than LGP-TP on four targets with $\text{p-value}<0.0162$ and $\text{p-value}<3e-4$, respectively, indicating significant superiority of LGP-TP over its opponents.
In addition, the mean rank of LGP-TP (5.45) has a large gap compared with the second-and third-highest mean ranks (i.e., FishCNN (4.65) and XGB (4.6)).
These results highlight the good overall performance of LGP-TP for predicting different kinds of biomass based on fish spectroscopic data.

Fig. \ref{fig:testR2}c shows the results of our ablation study. This paper introduces three kinds of tunable primitives into basic LGP. To verify the effectiveness of these three kinds of tunable primitives, we compare LGP-TP with basic LGP (i.e., without tunable primitives) and two crippled variants of LGP-TP. The two crippled variants only use tunable terminals only or use both tunable terminals and functions but without the multivariate linear regression function. The two crippled variants are denoted as LGP-TT and LGP-TTF, respectively. The area on the radar chart from basic LGP to LGP-TP grows consistently. The legend in Fig. \ref{fig:testR2}c shows the statistical results of Friedman's and Wilcoxon rank-sum tests, whose settings are the same as in Fig. \ref{fig:testR2}b. The overall p-value of Friedman's test of the ablation study is 0.0062, indicating a significant difference in the ablation study. Fig. \ref{fig:testR2}c shows growing mean ranks from LGP-TT to LGP-TP. Particularly, LGP-TP significantly improves the overall performance of basic LGP with a $\text{p-value}=0.019$.
The tunable primitives help LGP perform significantly better in more biomass targets based on the Wilcoxon rank-sum test, i.e., LGP-TT, LGP-TTF, and LGP-TP perform significantly better than basic LGP on 4, 5, and 6 biomass targets, respectively, with $\text{p-value}<0.001$.
The ablation study confirms that the proposed tunable primitives enhance the LGP performance for fish biomass prediction. 

\subsection{Interpretability Analysis}
\begin{figure}
    \centering
    \includegraphics[scale=0.71, viewport=15 170 570 830, clip=true]{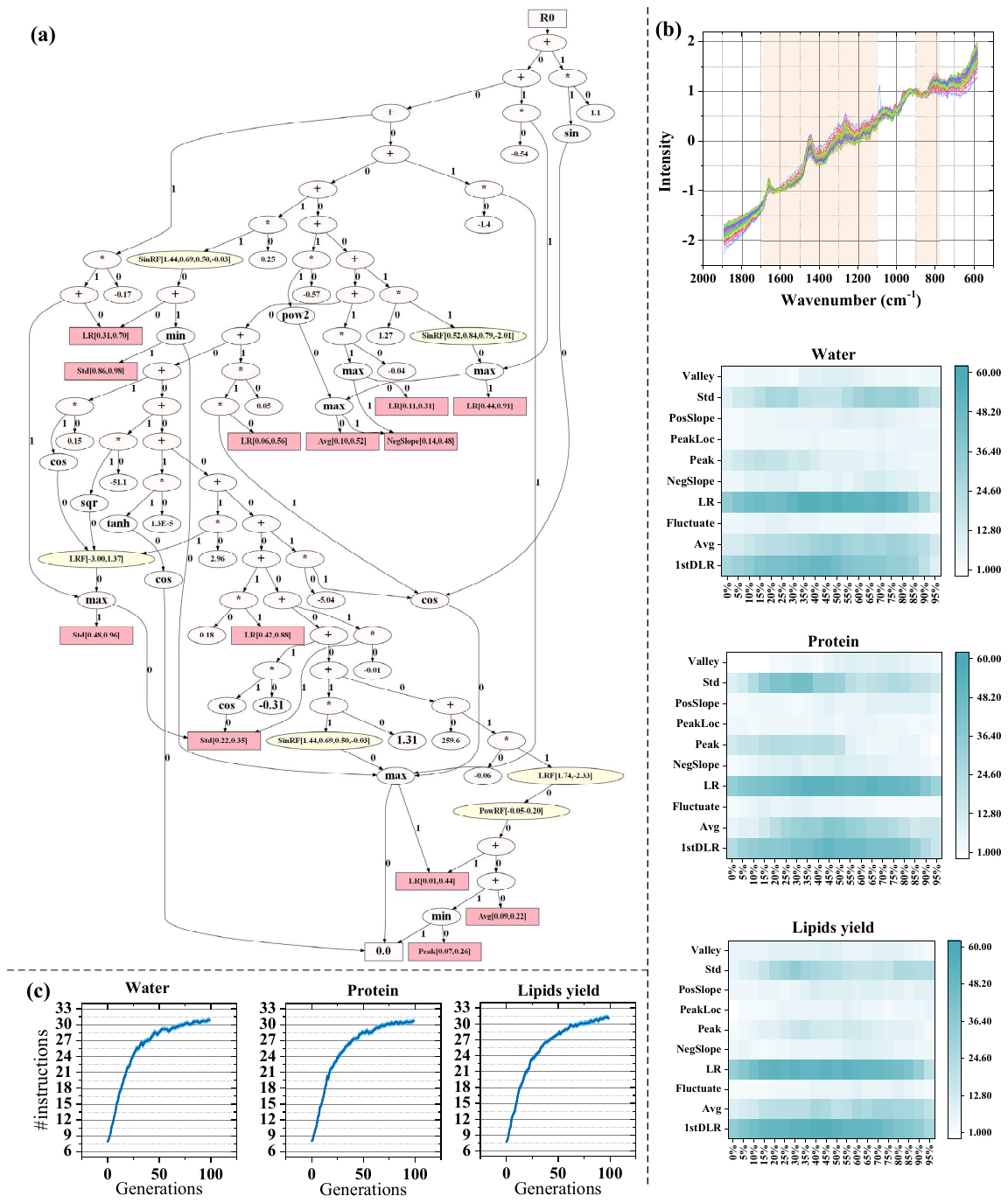}
    \caption{Interpretability analysis of LGP-TP. \textbf{(a)} An example synthesized regression model by LGP-TP for predicting water in fish meat, outputting results from the first register R0. \textbf{(b)} Highlighted input features by the result models. The upper part of sub-figure b is the Raman spectra data with highlighted features in yellow. 
    The three heat maps in the lower part show the frequency of input terminals over feature ranges (X-axis: percentage over the range of wavenumber, Y-axis: tunable terminals). The dark color indicates a high frequency over the 10 six-fold cross-validation. \textbf{(c)} Mean program size ($\pm$ std.) of synthesized regression models over generations from 10 six-fold cross-validation. We denote the number of effective instructions as the program size.}
    \label{fig:exampleprog}
\end{figure}

LGP-TP assists human in understanding spectral data by synthesizing symbolic regression models. The symbolic regression model allows us to infer and reason about the behaviors of the regression model. 
Fig. \ref{fig:exampleprog}a shows the directed acyclic graph of an example regression model for predicting water in fish meat. An amplified version of Fig. \ref{fig:exampleprog} refers to the supplement. The tunable terminals and functions are highlighted in pink and yellow, respectively. 
The graph nodes also show the input ranges of tunable terminals and the parameters of tunable functions within the square brackets. 
The multivariate linear regression function is split into ``$\times$'', ``$+$'', and coefficients in the top of the graph. The regression model contains a larger number of tunable terminals and functions, such as ``\texttt{LR}'' and ``\texttt{SinRF}'', implying an important role of tunable primitives in the regression model. Particularly, the example regression model prefers tunable terminals like ``\texttt{LR}'', ``\texttt{Std}'', and ``\texttt{Avg}''. These three kinds of features might have a high correlation with the water measurement. The example regression model prefers tunable functions like ``\texttt{SinRF}'' and ``\texttt{LRF}'' as well.
Regarding the multivariate linear regression function, the example model contains a large number of negative coefficients of multiplication, implying a negative correlation between the wavenumber features and the water measurement.

The synthesized regression models highlight some interesting characteristics of fish spectroscopic data. Fig. \ref{fig:exampleprog}b shows the highlighted features and the frequency of tunable terminals. Specifically, the lower part of Fig. \ref{fig:exampleprog}b shows the frequency heat maps of tunable terminals over the 10 six-fold cross-validation for predicting three example biomass targets. The X-axis of the heat maps is the percentage over the wavenumber range. We select three biomass targets, water, protein, and lipids yield, as examples to investigate the highlighted features. The three biomass targets are primary biomass in fish meat and usually account for more than 90\% of the total weight of a fish. We can see that these three biomass targets share a similar pattern of frequency heat maps. ``\texttt{LR}'', ``\texttt{1stDLR}'', ``\texttt{Avg}'', ``\texttt{Std}'', and ``\texttt{Peak}'' are the five most frequently used tunable terminals. Particularly, ``\texttt{LR}'' and ``\texttt{1stDLR}'' make use of nearly the whole range of wavenumber, while ``\texttt{Std}'' and ``\texttt{Avg}'' only prefer roughly 15$\sim$50\% and 75$\sim$85\%. The upper part of Fig.~\ref{fig:exampleprog} summarizes these overlapped ranges over the spectrum. We can see that the tunable primitives mainly focus on the first three and the last peaks over the spectroscopic data, implying a high correlation of these peaks with fish biomass.

The synthesized models are compact for interpreting. Fig. \ref{fig:exampleprog}c shows the average program size (and standard deviation) of the synthesized models, represented by the number of effective instructions in LGP programs. In the course of LGP-TP evolution, the program size grows from about 5 to 30 instructions in the three example biomass targets. It confirms that LGP-TP consistently synthesizes effective and compact programs.
In addition, the synthesized regression model is less than 30KB on average, which is acceptable for human analyzing. 

\subsection{Ability in Avoiding Overfitting}

The limited data quality and quantity easily lead to the overfitting issue in machine learning methods. LGP-TP relieves the overfitting by synthesizing necessarily competitive regression models.
To verify the ability of avoiding overfitting in LGP-TP, Fig.~\ref{fig:treatments}a shows the training and test performance over generations of LGP-TP in four example biomass targets. The training and test performance of LGP-TP in all the ten biomass targets refer to the supplement. Specifically, the orange curves indicate the training R$^2$ of LGP-TP, and the red curves indicate its test R$^2$. 
% The dashed lines with different colors indicate the training R$^2$ of other compared methods. 
We can see that the training R$^2$ of LGP-TP consistently rises over its evolution in all the four biomass targets, and the test R$^2$ rises or maintains at a certain level. We reckon that there is no significant overfitting over LGP-TP evolution since the gap between training and test R$^2$ does not widen. 
% By contrast, although the other six machine learning methods have a similar training performance to LGP-TP, their test performance imply the overfitting issue in some cases.
% On average, the LGP-TP performs competitively with other methods in terms of training performance. All the seven compared methods achieve a training R$^2$ of nearly 0.9.
% In contrast, the LGP-TP test R$^2$ does not significantly improve in some measurements, such as Ash and Fat. Although there is no a clear pattern of overfitting over LGP evolution (i.e., the improvement of training R$^2$ goes with the decrement of test R$^2$), 
% The test R$^2$ of LGP-TP in some biomass targets is far worse than the training R$^2$, implying that the training and test data are insufficient to represent the whole real relationship between fish spectroscopic data and biomass targets.

\subsection{Generality across Spectral Data Treatments}
\begin{figure}[t]
    \centering
    \includegraphics[scale=0.65, viewport=15 15 600 525, clip=true]{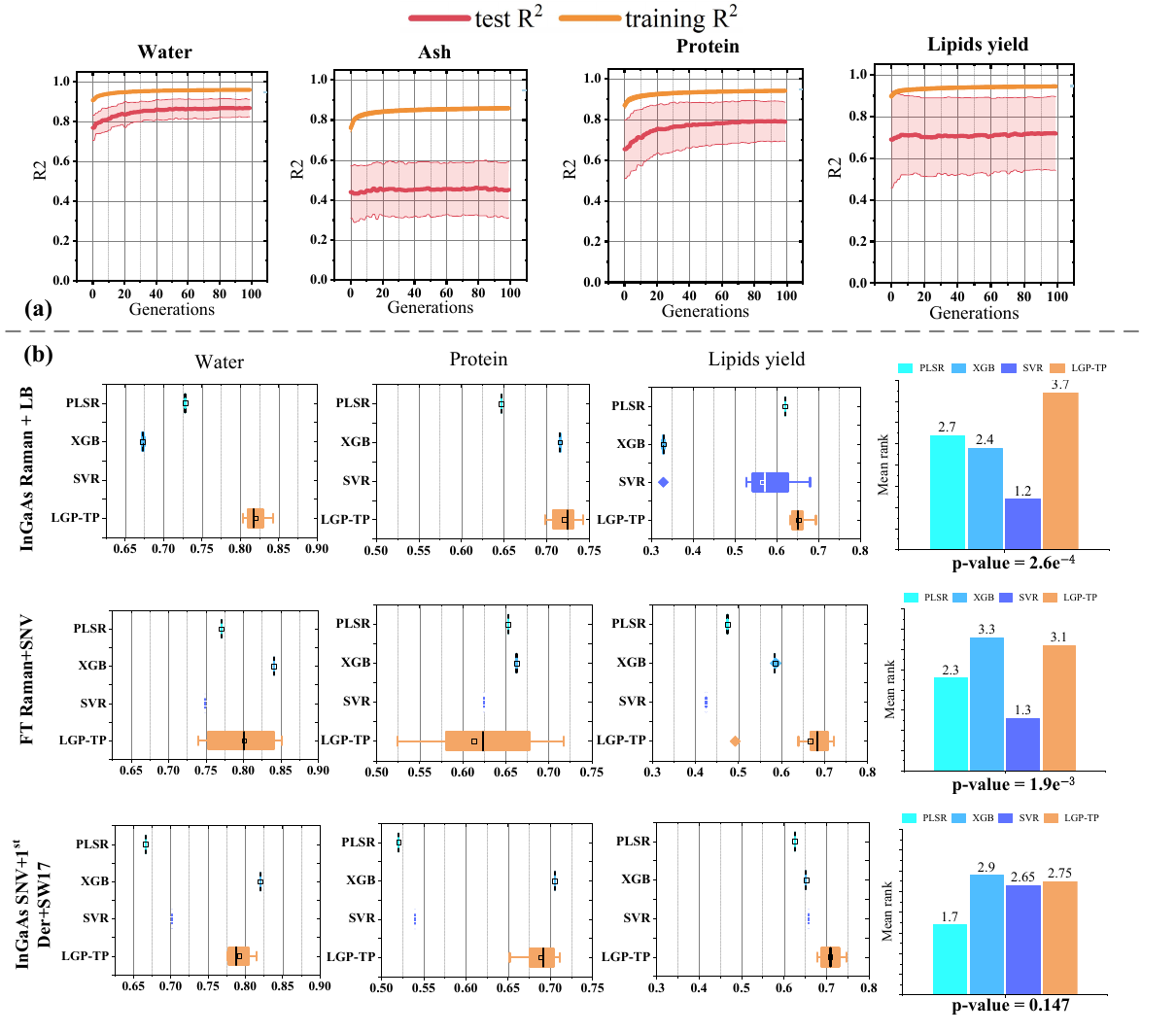}
    \caption{Training and test performance of LGP-TP and the general effectiveness across spectral data treatments. \textbf{(a)} The training and test R$^2$ of LGP-TP in four example biomass targets. \textbf{(b)} Test R$^2$ (X-axis) of the four compared methods for different data treatments over 10 six-fold cross-validation. The three rows stand for three different spectral data treatments. We show the box plots of test R$^2$ on three example biomass targets and the bar chart of mean ranks (and corresponding p-values) over the ten biomass targets by Friedman's test.}
    \label{fig:treatments}
\end{figure}

The proposed LGP-TP has a good generality to other data treatments of spectral data. 
The good generality of data treatments is important in practice as production lines might only handle certain data treatments other than the one used in our experiments.
This section verifies the generality of LGP-TP over three more data treatment techniques, 1) InGaAs Raman + LB, 2) FT Raman + SNV, and 3) InGaAs SNV + 1$^{st}$ Der + SW17. Specifically, 
\begin{enumerate}
    \item ``InGaAs Raman + LB'' applies linear baseline to process the InGaAs Raman at 1064nm with the wavenumber range of 580.109 to 202.533 (cm$^{-1}$);
    \item ``FT Raman + SNV'' indicates Fourier Transform (FT) Raman processed by SNV and with the wavenumber range of 4001.81 - 202.533 (cm$^{-1}$);
    \item ``InGaAs SNV + 1$^{st}$ Der + SW17'' indicates the InGaAs Raman at 1064num with the wavenumber range of 580.109 to 202.533 (cm$^{-1}$), processed by SNV, transformed into their 1$^{st}$-order derivative, and smoothened by a sliding window with size of 17 units.
\end{enumerate}
These data treatments lead to different wavenumber ranges and shapes of spectral data.
All these data treatments are split into six folds for cross-validation and augmented by the method mentioned in Section \ref{sec2:dataset}.

We verify the performance of LGP-TP via the comparison with PLSR, XGB, and support vector regression (SVR) \cite{ahmmed_detection_2023}. We select the water, protein, lipids yield, as example biomass targets to show the test R$^2$ distribution of the compared methods, as shown in Fig. \ref{fig:treatments}. Fig. \ref{fig:treatments} also shows the mean ranks by Friedman's test of the compared methods over all the ten biomass targets. LGP-TP has a very competitive test performance compared with the other methods. It achieves the best or second-best mean performance in most cases. The mean ranks by Friedman's test over the ten biomass targets confirm this observation. LGP-TP has the best or second-best mean ranks in different data treatments. Particularly, the second-best mean ranks in ``FT Raman + SNV'' and ``InGaAs SNV + 1$^{st}$ Der + SW17'' are very close to the best mean rank.
On the contrary, the performance of the other methods varies among the three data treatments. For example, XGB ranks third in ``InGaAs Raman + LB'', SVR ranks worst in ``InGaAs Raman + LB'' and ``FT Raman + SNV'', and PLSR rank worst in ``InGaAs SNV + 1$^{st}$ Der + SW17''.
The results imply the good generality of LGP-TP to other spectral data treatments.

\subsection{Generality across Regression Problems}
\begin{figure}[t]
    \centering
    \includegraphics[scale=0.75, viewport=20 20 500 480, clip=true]{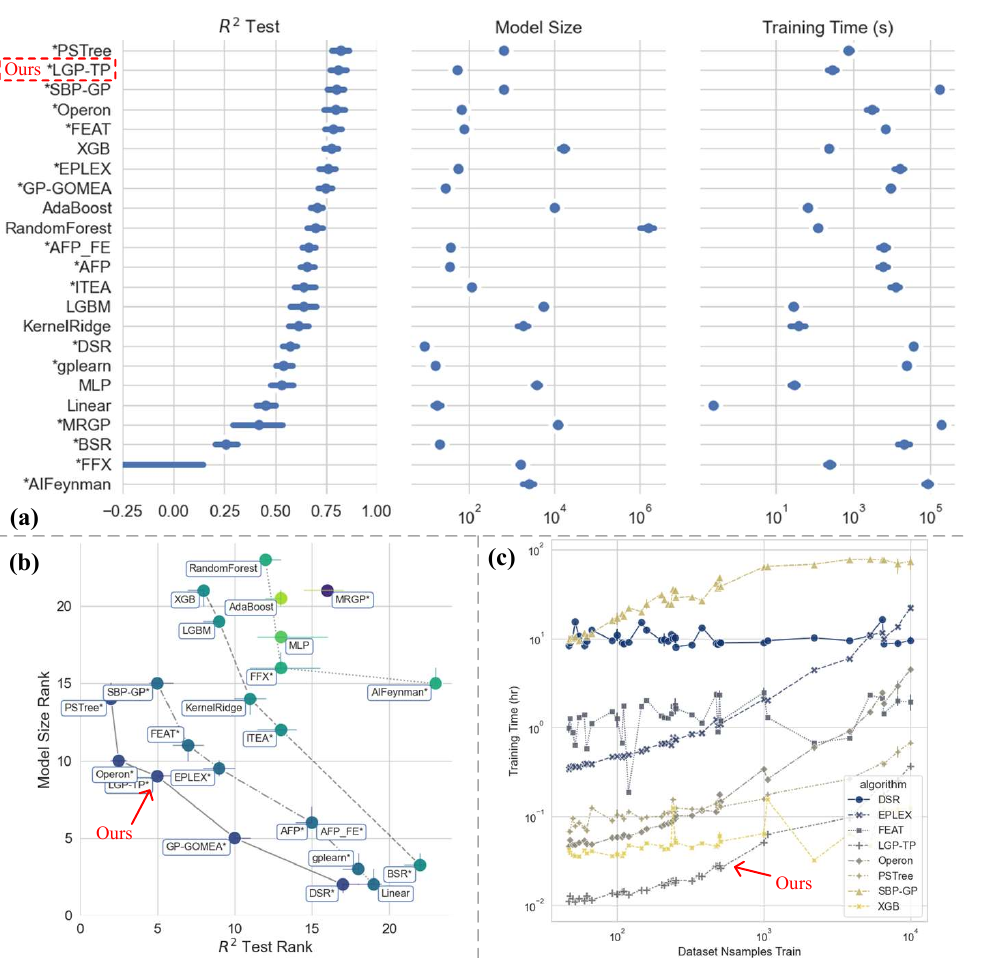}
    \caption{Results of SRBench. \textbf{(a)} Comprehensive comparison of benchmark methods in terms of test R$^2$, model size, and training time. \textbf{(b)} Non-dominate sets over model size rank and test R$^2$ rank. \textbf{(c)} Training time of benchmark methods over different numbers of training instances.}
    \label{fig:srbench}
\end{figure}

The proposed LGP-TP has a promising effectiveness across regression problems. 
Although this paper focuses on the fish spectroscopic dataset, we wonder if LGP-TP still has promising performance in other spectroscopic datasets and even a wider range of applications.
With this in mind, we extend LGP-TP to a popular symbolic regression benchmark, SRBench \cite{de_franca_srbench_2024}, to comprehensively investigate the potential generality of the proposed LGP-TP. 
Specifically, we apply LGP-TP to solve the black-box regression problems in SRBench, a total of 122 problems. These regression problems are mostly collected in practice, covering various application backgrounds and different levels of difficulties. Given that most of the problems in SRBench are with tabular data, we remove the spectroscopic-data-specific tunable terminals (e.g., ``\texttt{1stDLR}'' and ``\texttt{Std}'') and only retain ``\texttt{LR}'' as tunable terminals. The other tunable functions are kept. The primitive set of LGP-TP for solving SRBench includes the raw input features as well. LGP-TP for SRBench evolves 200 generations with a population size of 500 individuals. Each LGP-TP individual has up to 50 instructions and 8 available registers. The multi-variable linear regression function accepts the first four registers as inputs and stores the results in the output register R0.
Our experiments are implemented by the Java library ECJ \cite{luke_ecj_2017} and run on AMD EPYC 7702 computing nodes with a single thread, each thread having 16GB RAM.

% SBP-GP \cite{pawlak_semantic_2015}
We compare LGP-TP with 23 benchmark methods, including PSTree \cite{zhang_ps-tree_2022}, DSR \cite{petersen_deep_2021}, and many other popular regression methods (e.g., XGB and MLP). Fig. \ref{fig:srbench}a shows the main results of SRBench. LGP-TP has the second-best test R$^2$ amongst the benchmark methods and is very competitive with PSTree. In addition, LGP-TP approximately has the smallest model size and training time among the first six methods with the highest test R$^2$. Here, we multiply the number of effective instructions in an LGP program by a factor of 2.0 as the model size since functions in the primitive set of LGP-TP have up to two inputs. This implies that LGP-TP synthesizes concise yet very effective regression models with a short training time. LGP-TP likely has a good generality across various domains. 

Fig.~\ref{fig:srbench}b shows the non-dominated sets for optimizing two minimization objectives: model size rank and test R$^2$ rank. The proposed LGP-TP is in the middle position of the non-dominated set over all the benchmark methods, implying a good balance of LGP-TP between the two conflicting objectives, model size and model effectiveness. Fig.~\ref{fig:srbench}c shows the training time increment with the number of training instances. We can see that LGP-TP has the shortest training time in most cases, and the training time increases slower than many other benchmark methods. This implies a good feasibility for applying LGP-TP to larger datasets.

\section{Discussion}\label{sec12}

% restate your hypothesis or research question, restate your major findings, explain the relevance and the added value of your work, highlight any limitations of your study, describe future directions for research and recommendations. 

% Discussions should be brief and focused. In some disciplines use of Discussion or `Conclusion' is interchangeable. It is not mandatory to use both. Some journals prefer a section `Results and Discussion' followed by a section `Conclusion'. Please refer to Journal-level guidance for any specific requirements. 

% summarize the results, remove the discussion content in the last section?  highlighting three contributions 1) significantly improve the performance of PLSR and have a good interpretability 2) good generality in spectral data analysis  3) good generality in symbolic regression (many other domains).

This paper proposes a simple and effective LGP method for analyzing fish spectral data. We introduce three kinds of tunable primitives into the evolutionary framework of basic LGP and get LGP-TP.
% We take fish biomass composition prediction as a case study to verify the proposed LGP-TP. 
There are three major contributions in this paper. 

First, the results show that LGP-TP improves the overall performance of predicting fish biomass and achieves significantly better test performance than the compared methods, including one state-of-the-art method, on approximately half of the biomass targets. Furthermore, the synthesized regression models by LGP-TP have good interpretability. We can not only understand the model behaviors based on the symbolic representation but can also highlight the informative regions of spectroscopic data. This allows human researchers to summarize the new knowledge from the results. 
The comparison among different data treatments also shows that InGaAs Raman at 1064nm with the wavenumber range of 580.109 to 202.533 (cm$^{-1}$) and SNV correction is an effective way to highlight biomass information in fish spectral data.
% Although in some measurements, LGP-TP has an insignificant improvement of test performance after the training, that is because there are too few materials of these biomass measurements in fish meat, 

% for fish biomass composition prediction...

% the insignificant improvement of test performance: there are too few materials in fish meat such as ash. In addition, we only scan the fish meat for a short time, resulting in unstable spectroscopic features for these measurements.

% Second, our extension of LGP-TP to other spectral data treatments implies a good generality of the proposed LGP-TP in spectral data analysis. The good generality allows LGP-TP to work with various data treatment techniques in practice and has promising performance. This also saves a large amount of human effort and computation resources to try combinations between spectral data treatments and analyzing methods.
% Although the test performance of  over these three data treatments is not as good as the ones in Section \ref{sec2:testperformance}, 

% Third, our extension of LGP-TP to a popular symbolic regression benchmark, SRBench, verifies the promising performance of LGP-TP across a wide range of problems. LGP-TP can synthesize concise and effective regression models with a short training time. The results imply that tunable primitives are effective components for improving the LGP performance of synthesizing promising regression models. 

Second, our extension of LGP-TP to other spectral data treatments and a popular symbolic regression benchmark, SRBench, implies a promising performance of LGP-TP across a wide range of problems. The good generality allows LGP-TP to work with various spectral data treatments in practice and synthesize concise and effective regression models for various problems.

Third, our study shows that synthesizing regression models by genetic programming methods with tunable primitives is an effective strategy to solve complex regression problems. Specifically, LGP-TP searches symbolic structures of regression models by applying the basic genetic operators of GP, which has developed profoundly and has shown to be effective. On the other hand, LGP-TP tunes the coefficients in the regression models (i.e., the tunable components) by approximating the target outputs based on their immediate inputs. This strategy effectively finds concise symbolic solutions with appropriate tunable parameters and avoids the cumbersome error propagation from the end of a program to its beginning.

In the future, we plan to design more effective tunable primitives for LGP and apply them to more different domains. As the proposed tunable primitives in this paper only consider the linear approximation, designing tunable primitives with better learning ability is a potential direction. In addition, developing more advanced genetic operators to fine-tune both the symbolic combinations and their parameters for elite LGP individuals might be another direction to improve performance.

\section{Method}
\subsection{Overview}
% \begin{figure}
%     \centering
%     \includegraphics[scale=0.8, viewport=80 20 550 350, clip=true]{overview.pdf}
%     \caption{Overview of LGP-TP. The left yellow part is the evolutionary framework of LGP-TP. The right blue part details the production of the next-generation population. The main differences from the basic LGP evolutionary framework are highlighted in orange.}
%     \label{fig:overview}
% \end{figure}
% LGP-TP mostly follows the generational evolutionary framework of basic GP methods. Fig. \ref{fig:overview} shows the overview of LGP-TP. First, LGP-TP randomly initializes a population of individuals based on the given tunable and basic primitives. These primitives and the LGP representation define our search space. All the individuals are evaluated to predict the given target outputs. The prediction performance is the fitness of individuals. LGP searches regression models by varying on-hand competitive individuals. We select individuals with better fitness as parents and apply four genetic operators (i.e., micro and macro mutation, crossover, and swap) to produce offspring. The offspring production is equivalent to searching the neighborhood around good solutions in the search space, expecting to reach better fitness. By iteratively varying and keeping better individuals in the LGP population, it is likely for LGP to search competitive regression models.

\begin{figure*}
    \centering
    \includegraphics[scale=0.55, viewport=10 20 640 350, clip=true]{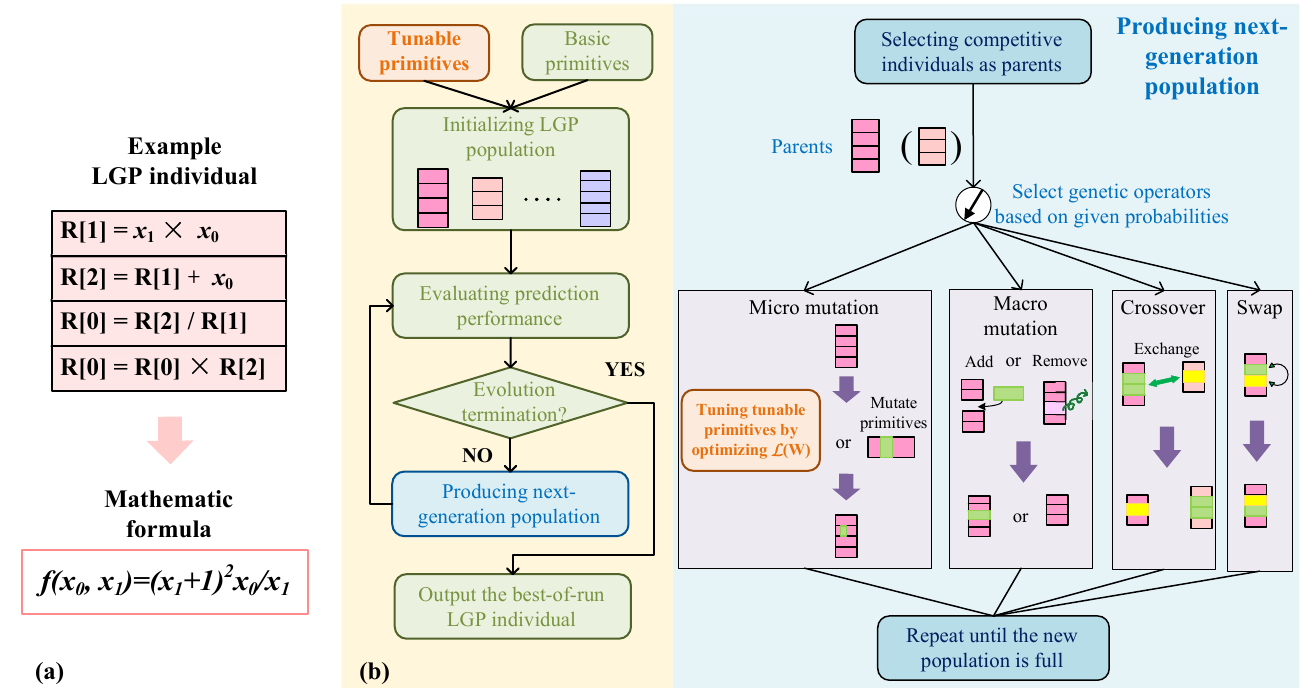}
    \caption{Overview of LGP-TP. \textbf{(a)} shows an example LGP program, representing a formula of $\frac{(x_1+1)^2x_0}{x_1}$. \textbf{(b)} is the training process of LGP-TP. The left yellow part is the evolutionary framework of LGP-TP. The right blue part details the production of the next-generation population. The main differences from the basic LGP evolutionary framework are highlighted in orange (i.e., tunable primitives and tuning their parameters).  }
    \label{fig:overview}
\end{figure*}

% The training process of LGP-TP mostly follows the generational evolutionary framework of basic GP methods. 
Fig. \ref{fig:overview} is the overview of LGP-TP.
An LGP individual has a linear representation of a sequence of register-based instructions. Fig. \ref{fig:overview}a shows an example LGP individual with four instructions. The LGP program initializes the registers (i.e., R[0] to R[2] in the example) as a certain value, such as 0 in our work. The LGP program executes the instructions sequentially. Each instruction accepts the values of source registers on the right and store the calculation results to the destination register on the left. The final output is stored in an output register (e.g., the first register R0 in this work). The example individual represents a mathematic formula of $\frac{(x_1+1)^2x_0}{x_1}$.

Fig. \ref{fig:overview}b shows the evolutionary framework of LGP-TP. First, LGP-TP randomly initializes a population of individuals based on the given tunable and basic primitives. These primitives and the LGP representation define our search space. All the individuals are evaluated to predict the given target outputs. The prediction performance of biomass is the fitness of individuals. 
LGP produces the next-generation population based on the individuals with competitive fitness by four genetic operators (i.e., micro and macro mutation, crossover, and swap).
% LGP searches regression models by varying on-hand competitive individuals. We select individuals with better fitness as parents and 
% apply four genetic operators (i.e., micro and macro mutation, crossover, and swap) to produce offspring. 

Macro mutation produces new programs by adding a new instruction into the LGP program or removing an instruction from the program. The term ``macro'' indicates that this mutation will change the total number of instructions in a program. In contrast, the micro mutation does not change the total number of instructions. It produces new programs by varying the primitives within an LGP instruction. 
Both swapping and crossover produce new programs by moving instructions. Specifically, The swapping operator swaps two consecutive instructions within an LGP program, while crossover exchanges two instruction sub-sequences from two different LGP programs \cite{Banzhaf1998}.

Particularly, we tune the tunable primitives in the micro mutation by minimizing the loss function $\mathcal{L}(\mathbf{W})$. 
As the multivariate linear regression function is always appended at the end of a program, it can be tuned after executing the whole new program.
% The offspring production is equivalent to searching the neighborhood around good solutions in the search space, expecting to reach better fitness. 
By iteratively varying and keeping better individuals in the LGP population, it is likely for LGP to search competitive regression models.

There are two major differences between LGP-TP and basic LGP, the primitive set and the genetic operator. Specifically, LGP-TP includes three kinds of tunable primitives in its primitive set and tunes the parameters of these tunable primitives in the genetic operator. The tunable primitives record the immediate inputs for themselves during program execution. The genetic operator produces new programs by optimizing the parameters of the tunable primitives, aiming to approximate the target output based on the immediate inputs. 

\subsection{Tunable Primitives}
Primitives are the elements to construct possible instructions in LGP. The primitives in LGP include terminals (e.g., problem inputs and registers) and functions (e.g., arithmetic operations). Although some existing studies of GP have included numerical numbers in their primitive set \cite{rovito_discovering_2023,Koza1992,topchy_faster_2001}, many of them just randomly generate numerical numbers without further tuning and cannot effectively evolve concise regression models with numerical numbers.

This paper proposes evolving tunable primitives in LGP. The main idea of tuning the parameters in the tunable primitives is to approximate the target outputs based on the immediate inputs of the LGP instruction.
We denote tunable primitives as $\mathcal{T}(\mathbf{X}, \mathbf{W}, \cdot)$, which is essentially a set of functions accepting the immediate input of the LGP instruction $\mathbf{X}=[\mathbf{x}_1,\mathbf{x}_2,...,\mathbf{x}_n]^T$, the tunable coefficients $\mathbf{W}$, and other primitive-specific arguments. $\mathbf{X}$ and $\mathbf{W}$ have different dimensions for different primitives. The output of $\mathcal{T}(\mathbf{X}, \mathbf{W}, \cdot)$ is a vector $[\tau(\mathbf{x}_1,  \mathbf{W}, \cdot), \tau(\mathbf{x}_2,  \mathbf{W}, \cdot),...,\tau(\mathbf{x}_n,  \mathbf{W}, \cdot)]^T \in \mathbb{R}^n$ where $\tau(\mathbf{x}_j, \mathbf{W}, \cdot)~j\in[1~..~n]$ is the element-wise result of $\mathcal{T}$, and $n$ is the number of training instances.
% where $\mathbf{X}\in \mathbb{R}^{n\times m}$ is a matrix representing the immediate input of the LGP instruction, each row for a data instance. $\mathbf{W}$ indicates a series of tunable parameters of the primitive $\tau$. 
We tune $\mathbf{W}$ to minimize the loss function 
\begin{equation}
  \mathcal{L}(\mathbf{W}) = ||\mathcal{T}(\mathbf{X}, \mathbf{W}, \cdot) - \mathbf{Y}||^2 = \sum_{j=1}^n(\tau(\mathbf{x}_j, \mathbf{W}, \cdot) - y_j)^2,  \label{eq:loss}
\end{equation}
where $\mathbf{Y}=(y_1, y_2,...,y_n)^T$ is the target output. There are three types of tunable primitives in our primitive set: tunable terminals, tunable functions, and a multivariate linear regression function. These three types of tunable primitives approximate the target output at the beginning, in the middle, and at the end of a program.

\subsubsection{Tunable Terminals}
We propose ten tunable primitives, including linear-regression-based terminal (\texttt{LR}), linear regression terminal on 1$^{st}$-order derivative (\texttt{1stDLR}), average terminal (\texttt{Avg}), standard deviation terminal (\texttt{Std}), fluctuation terminal (\texttt{Fluctuate}), negative-slope terminal (\texttt{NegSlope}), positive-slope terminal (\texttt{PosSlope}), peak terminal (\texttt{Peak}), valley terminal (\texttt{Valley}), and peak location terminal (\texttt{PeakLoc}). All these ten tunable primitives accept four arguments: the immediate input $\mathbf{X}$, coefficient vector $\mathbf{W}$, and two feature indices $\alpha$ and $\beta$. The immediate input of tunable terminals is the features of data instances. $\alpha$ and $\beta$ specify a range of consecutive input features $\mathbf{X}_{\alpha:\beta}$ for approximation. $\alpha$ and $\beta$ are searched as ephemeral random constants in GP \cite{Koza1992}. Specifically, we define the element-wise form $\tau(\mathbf{x}_j, \mathbf{W}, \cdot)$ of the ten tunable terminals as follows. 
% The inputs of the tunable terminals are always 
 
\texttt{LR} approximates the target outputs based on a range of input features. $\texttt{LR}(\mathbf{x}_j, \mathbf{W}_{L}, \alpha, \beta):\mathbb{R}^{\beta-\alpha+1}\times\mathbb{R}^{\beta-\alpha+2}\times\mathbb{R}\times\mathbb{R}\mapsto\mathbb{R}$ indicates that \texttt{LR} performs linear regression over the $\alpha^{th}$ to $\beta^{th}$ features $(\mathbf{x}_j)_{\alpha:\beta}$ with a vector of coefficients $\mathbf{W}_{L}$. Specifically, 
$$\texttt{LR}(\mathbf{x}_{j}, \mathbf{W}_{L}, \alpha, \beta) = [1, (\mathbf{x}_{j})^T_{\alpha:\beta}]\cdot \mathbf{W}_{L},$$ 
where $[1, (\mathbf{x}_{j})^T_{\alpha:\beta}]$ is a vector with an augmented constant of 1. We tune $\mathbf{W}_{L}$ to optimize Eq. (\ref{eq:loss}) via the linear least square regression method, as shown in Eq. (\ref{eq:LR}).
\begin{equation}\label{eq:LR}
    \hat{\mathbf{W}}_{L} = \left({\mathbf{X}_L}^T \mathbf{X}_L\right)^{-1}{\mathbf{X}_L}^T\mathbf{Y},
\end{equation}
where the first column of $\mathbf{X}_L := [1, \mathbf{X}_{:,\alpha:\beta}]$ is all ones.
% $$\mathbf{X}^{*}= \left[
% \begin{array}{c}
%      \left[1, \mathbf{X}_{1(\alpha, \beta)}\right]  \\
%      \left[1, \mathbf{X}_{2(\alpha, \beta)}\right]  \\
%     \colon \\
%      \left[1, \mathbf{X}_{n(\alpha, \beta)}\right]
% \end{array}
% \right]
% $$

$\texttt{1stDLR}(\mathbf{x}_{j}, \mathbf{W}_{D},\alpha, \beta):\mathbb{R}^{\beta-\alpha+1}\times\mathbb{R}^{\beta-\alpha+1}\times\mathbb{R}\times\mathbb{R}\mapsto\mathbb{R}$ approximates the target outputs by linear regression based on the first-order derivative of $(\mathbf{x}_j)_{\alpha:\beta}$. 
$$\texttt{1stDLR}(\mathbf{x}_{j}, \mathbf{W}_{D}, \alpha, \beta) = [1, \left[(\mathbf{x}_{j})_{\alpha+1:\beta} - (\mathbf{x}_{j})_{\alpha:\beta-1}\right]^T]\cdot \mathbf{W}_{D},$$ 
where $\left[(\mathbf{x}_{j})_{\alpha+1:\beta} - (\mathbf{x}_{j})_{\alpha:\beta-1}\right]^T$ indicates the first-order derivative of $(\mathbf{x}_j)_{\alpha:\beta}$. We tune $\mathbf{W}_D$ to approximate the target output $\mathbf{Y}$ via the linear least square regression method, as shown in Eq. (\ref{eq:1stD}) where $\mathbf{X}_D := [1,   \mathbf{X}_{:, \alpha+1:\beta} - \mathbf{X}_{:,\alpha:\beta-1}]$.
\begin{equation}
    \label{eq:1stD}
     \hat{\mathbf{W}}_{D} = \left({\mathbf{X}_D}^T \mathbf{X}_D\right)^{-1}{\mathbf{X}_D}^T\mathbf{Y}.
\end{equation}

The rest of the six tunable terminals follow the same form of 
$$\tau(\mathbf{x}_j, \mathbf{W}, \alpha, \beta) = \left[1, \gamma( (\mathbf{x}_{j})_{\alpha:\beta})\right]\mathbf{W}_a,$$
where $\gamma( (\mathbf{x}_{j})_{\alpha:\beta} ):\mathbb{R}^{\beta-\alpha+1}\mapsto\mathbb{R}$ represents a certain kind of information over the $\alpha^{th}$ to $\beta^{th}$ input features $(\mathbf{x}_j)_{\alpha:\beta}$. Specifically, $\gamma((\mathbf{x}_j)_{\alpha:\beta})$ can be:
\begin{enumerate}
    \item $\texttt{Avg}((\mathbf{x}_{j})_{\alpha:\beta})=\frac{1}{|\beta - \alpha+1|}\sum_{i=\alpha}^{\beta}(\mathbf{x}_{j})_i$ returns the average value of the $\alpha^{th}$ to $\beta^{th}$ input features. $(\mathbf{x}_{j})_i$ is the $i^{th}$ feature of input $\mathbf{x}_j$.
    \item $\texttt{Std}((\mathbf{x}_{j})_{\alpha:\beta})= \sqrt{ \frac{1}{|\beta - \alpha+1|}\sum_{i=\alpha}^{\beta} ((\mathbf{x}_{j})_i - \texttt{Avg}((\mathbf{x}_{j})_{\alpha:\beta}))^2 }$ returns the standard deviation of the $\alpha^{th}$ to $\beta^{th}$ input features.
    \item $\texttt{Fluctuate}((\mathbf{x}_{j})_{\alpha:\beta})=\frac{1}{|\beta - \alpha|}\sum_{i=\alpha+1}^{\beta} |(\mathbf{x}_{j})_i - (\mathbf{x}_{j})_i|$ returns the average accumulated variation over the $\alpha^{th}$ to $\beta^{th}$ input features.
    \item $\texttt{NegSlope}((\mathbf{x}_{j})_{\alpha:\beta})=\frac{1}{|\beta - \alpha|}\sum_{i=\alpha+1}^{\beta} \min\{(\mathbf{x}_{j})_i - (\mathbf{x}_{j})_{i-1}, 0\}$ and $\texttt{PosSlope}((\mathbf{x}_{j})_{\alpha:\beta})=\frac{1}{|\beta - \alpha|}\sum_{i=\alpha+1}^{\beta} \max\{(\mathbf{x}_{j})_i - (\mathbf{x}_{j})_{i-1}, 0\}$ return the average accumulated decrement and increment, respectively.
    \item  $\texttt{Peak}((\mathbf{x}_{j})_{\alpha:\beta})=\max_{i\in [\alpha..\beta]}~ (\mathbf{x}_{j})_i$ and $\texttt{Valley}((\mathbf{x}_{j})_{\alpha:\beta})=\min_{i\in [\alpha..\beta]}~ (\mathbf{x}_{j})_i$ return the maximum and minimum value over the $\alpha^{th}$ to $\beta^{th}$ input features, respectively.
    \item $\texttt{PeakLoc}((\mathbf{x}_{j})_{\alpha:\beta}) = \arg_{i\in [\alpha..\beta]} \max~ (\mathbf{x}_{j})_i$ returns the feature index of the maximum value over $\alpha$ to $\beta$.
\end{enumerate}
Let $\{\gamma((\mathbf{x}_j)_{\alpha:\beta})\}_{n}$ be an $n$-dimension vector, we have
${\Gamma} := [1, \{\gamma((\mathbf{x}_j)_{\alpha:\beta})\}_{n}] \in \mathbb{R}^{n\times 2}$ whose first column is all ones. We optimize Eq. (\ref{eq:loss}) by tuning $\mathbf{W}_a$ via the linear least square regression method. 
$$\hat{\mathbf{W}_a} =\left(\Gamma^T\Gamma\right)^{-1}\Gamma^T\mathbf{Y}.$$
To reduce the search space, the range $[\alpha.. \beta]$ covers up to 50\% of the total features for fish biomass prediction and up to 10\% of features for the SRBench problems.

\subsubsection{Tunable Functions}
We propose four tunable functions, including the linear-regression-based function (\texttt{LRF}), tunable trigonometric function (\texttt{SinRF}), tunable exponential function (\texttt{ExpoRF}), and tunable polynomial function (\texttt{PowRF}). 
All the four tunable functions are unary functions for the sake of simplicity, accepting one input value $x$ (the intermediate results of the LGP program). The immediate inputs for tunable functions are recorded during program execution.
To ensure the outputs of tunable functions better approximate the target outputs, the tunable functions approximate the residual error between the immediate input $x_j$ and the target output $y$.
We define the element-wise form $\tau(\mathbf{x}_j, \mathbf{W}, \cdot)$ of the four tunable functions as follows.
% We define the four tunable functions $\tau(\mathbf{X}_j, \mathbf{W})$ as follows. 
\begin{enumerate}
    \item $\texttt{LRF}(x_j, [\omega_0, \omega_1]^T)=\omega_0 + \omega_1 x_j$.
    \item $\texttt{SinRF}(x_j, [\omega_0, \omega_1, \omega_2, \omega_3]^T)=\omega_0 + \omega_1\sin(\omega_2x_j + \omega_3) + x_j$.
    \item $\texttt{ExpoRF}(x_j, [\omega_0, \omega_1]^T)=\omega_0 + (\omega_1^2 + 1)^{x_j}+x_j.$
    \item $\texttt{PowRF}(x_j, [\omega_0, \omega_1]^T)=\omega_0+|x_j|^{\omega_1} + x_j.$
\end{enumerate}
% We tune the parameters of LRF by the least square regression method and other tunable functions by gradient descent methods when optimizing Eq. (\ref{eq:loss}). 
Let $\{ \texttt{LRF}(x_j, [\omega_0, \omega_1]^T) \}_n$ be an $n$-dimension vector and $\mathbf{X}_F:= [1, \{ \texttt{LRF}(x_j, [\omega_0, \omega_1]^T) \}_n] \in \mathbb{R}^{n\times 2}$ be a matrix whose first column is all ones, we tune the parameters of \texttt{LRF} using the least square regression method.
$$[\hat \omega_0, \hat\omega_1]^T = \left( {\mathbf{X}_F}^T \mathbf{X}_F \right)^{-1} {\mathbf{X}_F}^T \mathbf{Y}.$$

We tune the parameters of the other three tunable functions (\texttt{SinRF}, \texttt{ExpoRF}, \texttt{PowRF}) by a gradient descent method with a step size $\lambda$ ($\lambda=0.1$ in our work) to optimize Eq. (\ref{eq:loss}). Specifically,
% For  whose parameters are optimized by gradient descent methods with a step size $\lambda$ ($\lambda=0.1$ in our work), 
$$\hat\omega = \omega - \lambda\frac{\partial \mathcal{L}}{\partial \omega} / ||\frac{\partial \mathcal{L}}{\partial \omega}||.$$
and have, for \texttt{sinRF}:
$$\frac{\partial \mathcal{L}}{\partial \omega_0} = \sum_j^n 2(\omega_0 + \omega_1\sin(\omega_2x_j + \omega_3) + x_j - y_j),$$
$$\frac{\partial \mathcal{L}}{\partial \omega_1} = \sum_j^n 2(\omega_0 + \omega_1\sin(\omega_2x_j + \omega_3) + x_j - y_j)\omega_1\sin(\omega_2x_j + \omega_3),$$
$$\frac{\partial \mathcal{L}}{\partial \omega_2} = \sum_j^n 2(\omega_0 + \omega_1\sin(\omega_2x_j + \omega_3) + x_j - y_j)\omega_1\cos(\omega_2x_j + \omega_3)x_j,$$
$$\frac{\partial \mathcal{L}}{\partial \omega_3} = \sum_j^n 2(\omega_0 + \omega_1\sin(\omega_2x_j + \omega_3) + x_j - y_j)\omega_1\cos(\omega_2x_j + \omega_3).$$

For \texttt{ExpoRF}:
$$\frac{\partial \mathcal{L}}{\partial \omega_0} = \sum_j^n 2(\omega_0 + (\omega_1^2 + 1)^{x_j} + x_j - y_j),$$
$$\frac{\partial \mathcal{L}}{\partial \omega_1} = \sum_j^n 2(\omega_0 + (\omega_1^2 + 1)^{x_j} + x_j - y_j)2\omega_1 x_j(\omega_1^2 + 1)^{x_j - 1}.$$

For \texttt{PowRF}:
$$\frac{\partial \mathcal{L}}{\partial \omega_0} = \sum_j^n 2(\omega_0+|x_j|^{\omega_1} + x_j - y_j),$$
$$\frac{\partial \mathcal{L}}{\partial \omega_1} = \sum_j^n 2(\omega_0+|x_j|^{\omega_1} + x_j - y_j)|x_j|^{\omega_1} \ln(|x_j|).$$

The parameters $\omega$ in the tunable functions have a limited value range of $[-3, 3]$.
\subsubsection{Multivariate linear regression function}
The proposed multivariate linear regression function (\texttt{MVLR}) is a special tunable function that is always appended at the end of LGP programs. It integrates the values from multiple registers into one output value (i.e., storing the result in an output register). To keep LGP programs concise, each LGP program only has one \texttt{MVLR} function for each output register. We define the element-wise form of \texttt{MVLR} as follows.
$$\texttt{MVLR}(\mathbf{x}_j, \mathbf{W}_M) : \mathbb{R}^{r}\times\mathbb{R}^{r+1}\mapsto \mathbb{R} =[1, \mathbf{x}^T_j]\cdot\mathbf{W}_{M},$$ 
where $r$ is the number of registers inputted to \texttt{MVLR}.
$\mathbf{W}_M$ is tuned based on the register values after executing the precedent instructions in the LGP program. 
\texttt{MVLR} accepts $r=20$ register values for fish spectroscopic data ($\mathbf{x}_j\in \mathbb{R}^{20}$) and accepts $r=4$ register values for SRBench problems ($\mathbf{x}_j\in \mathbb{R}^{4}$). We tune $\mathbf{W}_{M}$ by optimizing Eq. (\ref{eq:loss}) via the least square regression method where $\mathbf{X}_{M} := [1,\mathbf{X}]\in \mathbb{R}^{n\times (1+r)}$ with the first column of all ones. 
$$\hat{\mathbf{W}}_{M} = ({\mathbf{X}_{M}}^T {\mathbf{X}_{M}})^{-1}{\mathbf{X}_{M}}^T\mathbf{Y}.$$

\subsubsection{Overall Primitive Set}
The overall primitive set includes the proposed tunable primitives and basic arithmetic operations ($+$, $-$, $\times$, Analytic Quotient), trigonometric operations ($\sin$, $\cos$, $\tanh$), boolean operations ($\max$, $\min$), and others ($\sqrt x$, $x^2$, $\exp(x)$, and $\ln(|x|)$).
These primitives consist of all possible instructions together with registers. In this work, an LGP instruction includes one destination register, one function, and two terminals such as problem inputs and source registers.

\bibliography{sample.bib}

\section*{Author contributions statement}

Z. H. proposed the algorithm in this manuscript, implemented the experiment, and wrote the draft. B. X. and M. Z. helped design the algorithm. J. S. R., K. C. G., and D. P. K. collected fish spectral data and designed the experiments. All authors checked and revised the manuscript.

\section*{Data Availability}
The fish spectrum data is not openly available due to commercial sensitivity. However, they are available from the data provider (Daniel.Killeen@plantandfood.co.nz) on reasonable request. The SRBench dataset is a public dataset, which can be accessed via https://cavalab.org/srbench/.

\section*{Competing interests}
The author(s) declare no competing interests.

% \section*{Additional information}

% To include, in this order: \textbf{Accession codes} (where applicable); \textbf{Competing interests} (mandatory statement). 

% The corresponding author is responsible for submitting a \href{http://www.nature.com/srep/policies/index.html#competing}{competing interests statement} on behalf of all authors of the paper. This statement must be included in the submitted article file.

% \begin{figure}[ht]
% \centering
% \includegraphics[width=\linewidth]{stream}
% \caption{Legend (350 words max). Example legend text.}
% \label{fig:stream}
% \end{figure}

% \begin{table}[ht]
% \centering
% \begin{tabular}{|l|l|l|}
% \hline
% Condition & n & p \\
% \hline
% A & 5 & 0.1 \\
% \hline
% B & 10 & 0.01 \\
% \hline
% \end{tabular}
% \caption{\label{tab:example}Legend (350 words max). Example legend text.}
% \end{table}

% Figures and tables can be referenced in LaTeX using the ref command, e.g. Figure \ref{fig:stream} and Table \ref{tab:example}.

\end{document}